# Classifying the Valence of Autobiographical Memories from fMRI Data

Alex Frid[2,3*], Larry M. Manevitz[1,2] and Norberto Eiji Nawa[4,5]

[1]Department of Computer Science, Ariel University, Ariel, Israel
[2]Neurocomputation Laboratory, Department of Computer Science, University of Haifa, Haifa, Israel
[3]The Laboratory of Clinical Neurophysiology, Technion Faculty of Medicine, Haifa, Israel
[4]Center for Information and Neural Networks (CiNet), National Institute of Information and Communications Technology (NICT), Japan
[5]Graduate School of Frontiers Biosciences, Osaka University, Japan

**\* Correspondence:**
Corresponding Author: Alex Frid
alex.frid@gmail.com

## Abstract

We show that fMRI analysis using machine learning tools are sufficient to distinguish valence (i.e., positive or negative) of freely retrieved autobiographical memories in a cross-participant setting. Our methodology uses feature selection (ReliefF) in combination with boosting methods, both applied directly to data represented in voxel space. In previous work using the same data set, Nawa and Ando showed that whole-brain based classification could achieve above-chance classification accuracy only when both training and testing data came from the same individual. In a cross-participant setting, classification results were not statistically significant. Additionally, on average the classification accuracy obtained when using ReliefF is substantially higher than previous results - 81% for the within-participant classification, and 62% for the cross-participant classification. Furthermore, since features are defined in voxel space, it is possible to show brain maps indicating the regions of that are most relevant in determining the results of the classification. Interestingly, the voxels that were selected using the proposed computational pipeline seem to be consistent with current neurophysiological theories regarding the brain regions actively involved in autobiographical memory processes.

## 1. Introduction

Memory encoding and retrieval are arguably two of the most complex cognitive processes performed by humans [1], [2]. Study of this process is a central concern of psychology and memory researchers. A technological window on cognitive activities in general, and memory in particular, is the use of neuroimaging techniques to help elucidate the neurophysiological basis underlying such memory processes. The combination of neuroimaging technology with machine learning techniques [3]–[6] has opened a promising front in the past decade or so. A common approach has been to attempt to identify cognitive processes, states or disorders from neuroimaging data using various types of machine learning techniques. Although progress has been made in the field, the subtleties of memory processes pose a considerably more challenging task, as compared to tasks involving, for instance, the perception of visual stimuli.

The retrieval of episodic memories derived from events experienced from one's personal past – i.e., autobiographical memories – has been shown to recruit a brain-wide network of regions, such as medial and lateral temporal structures, most notably the hippocampus (HC) and parahippocampus, prefrontal areas including dorsolateral and ventromedial regions, posterior midline regions such as precuneus (PCUN) and retrosplenial cortex (RSC), and lateral parietal cortex [7]–[10]. The retrieval of emotional memories, i.e., memories of events associated with greater levels of arousal or valence [11], in particular, has been associated with heightened activity in prefrontal regions, and oftentimes the amygdala [12]–[14].

In Nawa and Ando [15] the authors investigated freely retrieved autobiographical memory formation and showed that, in fact, it is possible to reliably distinguish between autobiographical memory retrieval and a completely different cognitive task (in their case, counting backwards) based on data from a single fMRI scan (or volume). This worked also in a cross-participant setting; i.e., when using data from n-1 participants to train the machine learning classifier and using data from the left-out participant to test the classifier generalizability and accuracy. They decided to pursue this without choosing *a priori* regions of interest (ROIs), under the assumption that since memory retrieval is a complex process involving many regions, there could be loss of information by focusing in selected brain areas. They also proceeded to the even more delicate task of distinguishing between two different kinds of autobiographical memory, those with positive valence from those with negative valence. They did succeed in this task at a significant level but only in a *within-participant* setting, i.e., when using a subset of the scans not used for training when testing the generalizability and accuracy but with all scans collected from the same individual. However, they did not succeed in doing this for generalizing to memories from a novel individual (the cross-participant setting).

In this work, we show that it is possible to dramatically increase the performance for the cross-participant setting when applying a combination of techniques, provided that a strong voxel selection is performed prior to the machine learning training stage proper. Although this may be in contradiction with the idea of using data from the entire brain because of the complexity of memory retrieval processes, feature selection may have an effect akin to strengthening the signal to noise in the training signal. Moreover, by examining the location of the voxels most often selected across participants, it was possible to confirm that qualitatively, some of the voxels driving the machine learning classification were located on brain regions that are known to be involved with autobiographical memory processes. This suggests that machine learning techniques may also serve as tools of discovery by suggesting to neuroscientists areas of potential interest with regard to a cognitive function, in an information theoretic principled way.

## 2. Methodology

### 2.1. Experimental Data

In this work we used the same data set reported in Nawa and Ando [15]. Their protocol for the autobiographical recall task was as follows (for a more detailed explanation of the experiment, see [15]). Participants performed three types of mental tasks while in the MRI scanner: a countdown task, a positive autobiographical memory retrieval task, and a negative autobiographical memory retrieval task. Figure 1 shows a graphical representation of the data collection protocol. In essence, each participant participated in 12 scanning sessions, each session consisted of three blocks of either "positive autobiographical memory retrieval and counting backwards tasks" or three "negative memory and then counting tasks". Each such experimental task block consisted of memory (either positive or negative) task and countdown task. The memory task lasted 32 seconds followed by 16 seconds of rest after which the countdown task (32sec) was executed. Each block type was repeated 3 times in each session. In 6 sessions, subjects switched between negative memory and countdown tasks (i.e., M_A and Cn tasks) and in the remaining 6 sessions they switched between tasks between positive memory and countdown tasks (i.e., M_B and Cn tasks). All sessions were performed in the same day. One hundred forty-seven scans were acquired in each session (TR = 2 s, 33 4-mm slices, in-plane spatial resolution of 3 mm x 3 mm). Each single scan was encoded as a vector containing the blood-oxygen level dependent (BOLD) signal of voxels covering the entire brain. From a machine learning perspective, the most important points of the study are that (i) there were no restrictions on the contents of the memories, i.e., the retrieved memories did not have to necessarily be associated with landmark personal events but rather with more mundane episodes; participants were only requested to thoroughly "relive" the original event during the scan, by focusing on the circumstances that led to the event and other details associated with it, and (ii) the relatively long scanning time that allowed the participants to "freely engage" into the execution of the task.

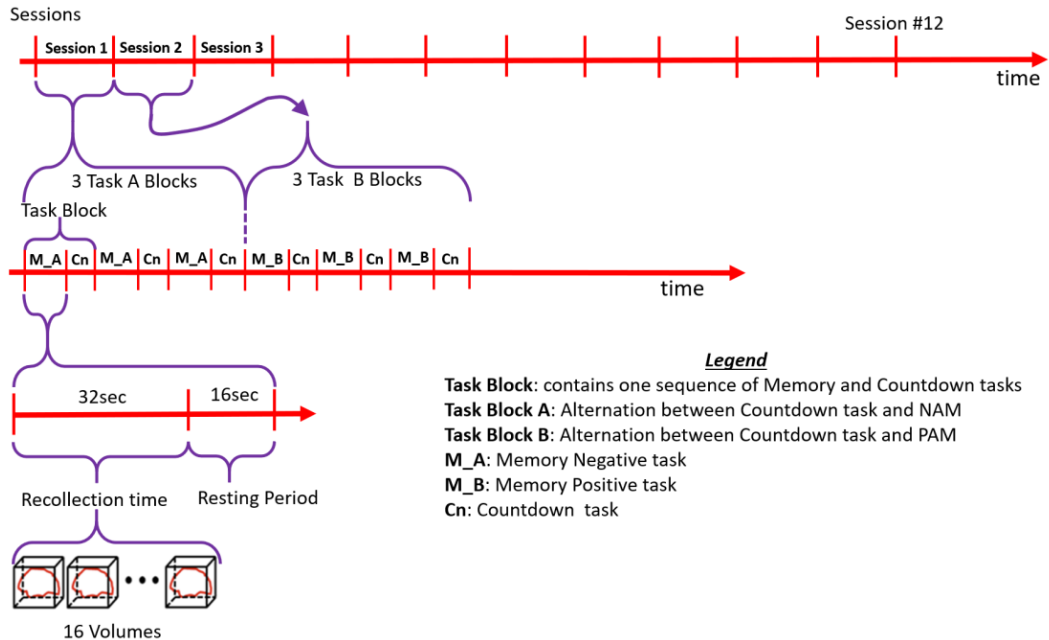

*Figure 1* – 12 Sessions were recorded from participant, each session consisted from three memory tasks and three counting tasks (in alternating fashion), each task consisted from 32 seconds of recording time and 12 seconds of resting period. The scanning speed was 2 seconds resulting in 16 volumes per task.

A machine learning-based scheme was then employed to predict the valence of the autobiographical memories recalled by human subjects based on the information contained in a single functional magnetic resonance imaging (fMRI) scan. Subjects (N = 11, 6 females, age 21 - 37, average 28.2 years old, right-handed) were asked beforehand to prepare a list of happy and sad events that they had experienced in the past. During scanning, subjects were asked to keep their eyes closed and given auditory cues which indicated whether they should alternate between counting down numbers and recollecting positive autobiographical memories or counting down and recollection of negative autobiographical memories (see Figure 1 for details). The memory recollection task was conducted in a self-paced manner, during which subjects were asked to remember as many details as possible about the events.

### 2.2. Machine Learning and Data Mining Methodology

The entire process of classification is depicted in Figure 2. After a standard data preprocessing step (for details, see [15]), the data of the participants were divided into training and testing groups. The machine learning procedures that were applied on the training group consisted of: (i) scoring the voxels by their relevance to the classification task, and then performing feature selection using a form of thresholding (ii) construction of a classification method using tree stumps from the selected voxels and (iii) a version of the AdaBoost methodology on the stump. Next, the classifier was evaluated on the test group data, and we examined both the efficiency of the selected features (i.e., brain regions), and the performance of the classification algorithm.

These steps are explained in more detail below.

### 2.2.1. Pre-processing step

Imaging data was acquired on a 3T Siemens Magnetom Trio, A Tim System scanner (Siemens Healthcare, Erlangen, Germany) equipped with a 12-channel standard head coil. A standard preprocessing pipeline was adopted before the data was used to train and test the classifiers, which included temporal slice time correction, spatial realignment, normalization to a standard stereotaxic space (Montreal Neurological Institute, MNI) and spatial smoothing (Gaussian kernel of 8-mm full width at half maximum). Those steps were performed using the functions available in SPM 5 (Wellcome Trust Centre for Neuroimaging, UK, http://www.fil.ion.ucl.ac.uk/spm/software/spm5). During the spatial normalization, for each subject, a visual inspection of the results was performed by validating the registration of selected landmark points with reference to the brain template. Most importantly, normalized images were rewritten using the voxel size originally employed during data collection (3 x 3 x 4 mm).

### 2.2.2. Feature selection

The goal of feature selection is twofold. First, from a purely data driven aspect, feature selection is an effective way to reduce the number of dimensions, and avoid the "curse of dimensionality" problem [16]. Since each voxel represents a feature, the number of possible features is much larger than the amount of data points, i.e., brain scans: there were nearly 40,000 voxels for a total of about 600 scans (from the two conditions of positive and negative autobiographical memories for each one of the participants). Not having to deal with such an extremely large number of features, given the small number of data points, is likely to be crucial to improve classification results. The second goal of feature selection has to do with the role of machine learning classification in realm of neuroscience research; by keeping the features in voxel space during the process of dimensionality reduction, and the machine learning based classification per se (as opposed to defining features in a more abstract space, e.g., the inner level representations common to deep learning networks [17]), it is possible to more easily understand the results, i.e., which areas of the brain contain the most discriminative information for the purposes of classification), and consequently, obtain more meaningful insights with regard to the brain regions or networks.

This led us to the design of the following "two staged" feature selection process: first, a "gross" feature selection method is applied in order to select a subset of the features to concentrate the computational power on. This step is explained in this section. Later, during the classification procedure, a finer and multivariate feature selection method is applied in order to focus on the most informative features and fuse them into a single spatial activation pattern. This step is explained in the next section (Classification).

For the first, "gross", feature selection step, a ReliefF [18] algorithm is used in this scheme. In essence, the algorithm works by randomly sampling instances and for each such instance locating its '$k$' nearest neighbours from the same and opposite classes[1]. The values of the features of the nearest neighbours are compared to the sampled instance and used to update (for each sampled instance) what is called the relevance scores for each feature (i.e., the closer a sample is to a same class sample, the higher its relevance). The relevance scores are calculated by the following equation:

---

[1] Note: Should all training data points be selected, the algorithm is deterministic. In this study, we sampled 10% of *the data.*

$$W_i = W_i - \frac{\sum_{k=1}^{K} Dist_H(k)}{n \cdot K} + \frac{\sum_{k=1}^{K} Dist_M(k)}{n \cdot K} \tag{1}$$

Here 'K' is the number of neighbors, 'n' is the number of repetitions (i.e., the number of randomly selected instances from the dataset), 'H' are the neighbors selected from the current sample's group (i.e., "Hits") and 'M' are the selected neighbors from the opposite group (i.e., "Misses"). The $Dist_H(k)$ is the distance between the selected instance and its k-th nearest neighbor in H (the same for M).

This algorithm has several properties that makes it suitable for this type of preliminary feature selection. First, the algorithm provides feature ranking, in terms of its relevance to the classification task. Second, since the features are evaluated in "K-nearest neighbor" (KNN) fashion, its probability of error is close to a Bayesian decision rule (see [19]) and strongly correlated with impurity functions (in other words it calculates the amount of probability of a specific feature that is classified incorrectly when selected randomly [20]). This in turn makes this process relatively resilient to noise that is typically observed in fMRI data.

The specific choice of "k" depends on a balance between the amount of data, computational resources and desired accuracy [21]. In essence, higher "K" will redure the sensitivity of algorithm to noise by calculating the average distance. On the other hand, in small datasets, higher "K" can account unrelated information. In our work, (see below) we settled on K=3. With substantially additional data and computational power, our impression from minor tests is that a larger K (e.g. K= 6) might improve our results.

Thus, following the above methodology results in a choice of the subset of voxels both by maximizing the certainty of decision individually on each voxel as well as being relatively resilient to noise in the training data.

However, since ReliefF is an iterative algorithm, in our use of the algorithm we have to consider that for each feature (40,000 voxels) all data points are needed, resulting in (~600 scans per participant x 11 participants x the value of K) iterations so the algorithm could be very time-consuming. In order to make this feasible and to reduce the influence of the noisy features we used a sampling methodology which we chose to do over the data. That is, not all the data-points were used in order to evaluate feature's influence.

After the relevance values are computed, we chose (i.e., threshold) the best 2500 voxels as our features, i.e., N = 2500 voxels. Since this feature selection is really a univariant method; while our classification algorithm (below) is multi-variate a balancing between introducing too much noise in the classification algorithm (from not sufficiently relevant voxels) and having sufficient information available for the classification algorithm is required. In addition we also needed to keep in mind the processing speed (i.e. training and validation times). We heuristically chose 2500 as approximately 5 -10% of the total voxels in the brain volume since (i) we expected that only a small fraction of the brain in engaged in the specific task and (ii) in analyses of other tasks [3], [4], [22] we found that this amount of voxels on similar resolution scans gives an appropriate compromise.

### 2.2.3. Classification

While the initial feature selection scheme performed by the ReliefF algorithm aimed at finding the subset of voxels with reduced levels of noise (as explained in the previous section), the classification step aims to find a multivariate activation pattern between those voxels that can be reliably associated in distinguishing two cognitive states, i.e., any two of remembering positive autobiographical memory, a negative autobiographical memory, or counting numbers backwards. In order to achieve that, an ensemble learning method was used as the classification scheme. More specifically, in this paper we used the AdaBoost [23], [24] method. In earlier work [15] Nawa used support vector machines (SVM) as the algorithm on the entire brain. Some advantages of the AdaBoost method (as opposed to SVM or neural networks) can be seen by considering Eq. 2 below. The underlying idea behind ensemble learning is to (i) find the best weak learners, i.e., maximally correlated with the desired classification, and then (ii) find the best linear combination of weak learners to strengthen the final classification. In our case this means finding the voxels whose ensembled activation is maximally correlated (even if the correlation have small positive or negative values) to the valence of autobiographical memories; and then finding the best linear combination of these voxels, in order to obtain a robust multivariate classification tool. Furthermore, since all the steps take place in voxel space, we can directly visualize, e.g., on a standard MNI brain, the location of the voxels, and their relative importance for classification accuracy. This is a relatively direct way to pick out important patterns of activation.

In other words, we can represent this process as finding the appropriate weights of each of the selected voxel-based classifiers. Thus, our classifier is of the form of:

$$P(x) = w_1 f_1(x_1) + w_2 f_2(x_2) + \ldots + w_i f_i(x_i) \qquad (2)$$

where the final decision is based on the sign of P. (Here $x_i$ is the i'th coordinate of $x$.)

One way to tackle this problem is to find the $w_i f_i(x_i)$, i.e., turn each voxel into a "weak classifier", $f_i(x_i)$, that is based on a single voxel and performs slightly better than chance level. The $w_i$ would be the weight of this classifier in the final decision. This can be achieved by using a decision tree stump methodology (i.e., a one-level decision tree) [25]. Practically, each voxel from the remaining set (i.e., after the feature selection procedure) splits the training dataset into two groups, namely positive and negative autobiographical memories, using a cut (in the voxel value) that gives the best separation gain. In this scheme, the gain was computed using Cross Entropy measure [26].

One method of finding the $w_i$'s (see Eq. 2) is to proceed in a greedy fashion by starting with just one voxel (the highest correlated one) and then, in each iteration for each remaining voxel separately, find the optimal w; and then compare the best results over all choices of the voxels. In essence, we would like to minimize the training error $E_i$ (see Eq. 3) with each iteration *i* (i.e. with each addition of a new classifier trained on additional modality). This iterative method is called AdaBoost [23].

$$E_i = \sum_i E[P_{i-1}(x_i) + w_i f_i(x_i)] \qquad (3)$$

Here $E_i$ is the sum of the training error at iteration $i$, $P_i$ is the final classifier after $i$ iterations, $f_i(x_i)$ is the output hypothesis produced by a single classifier for each voxel in the training data-set and $w_i$ is the weight assigned to the $i$'s.

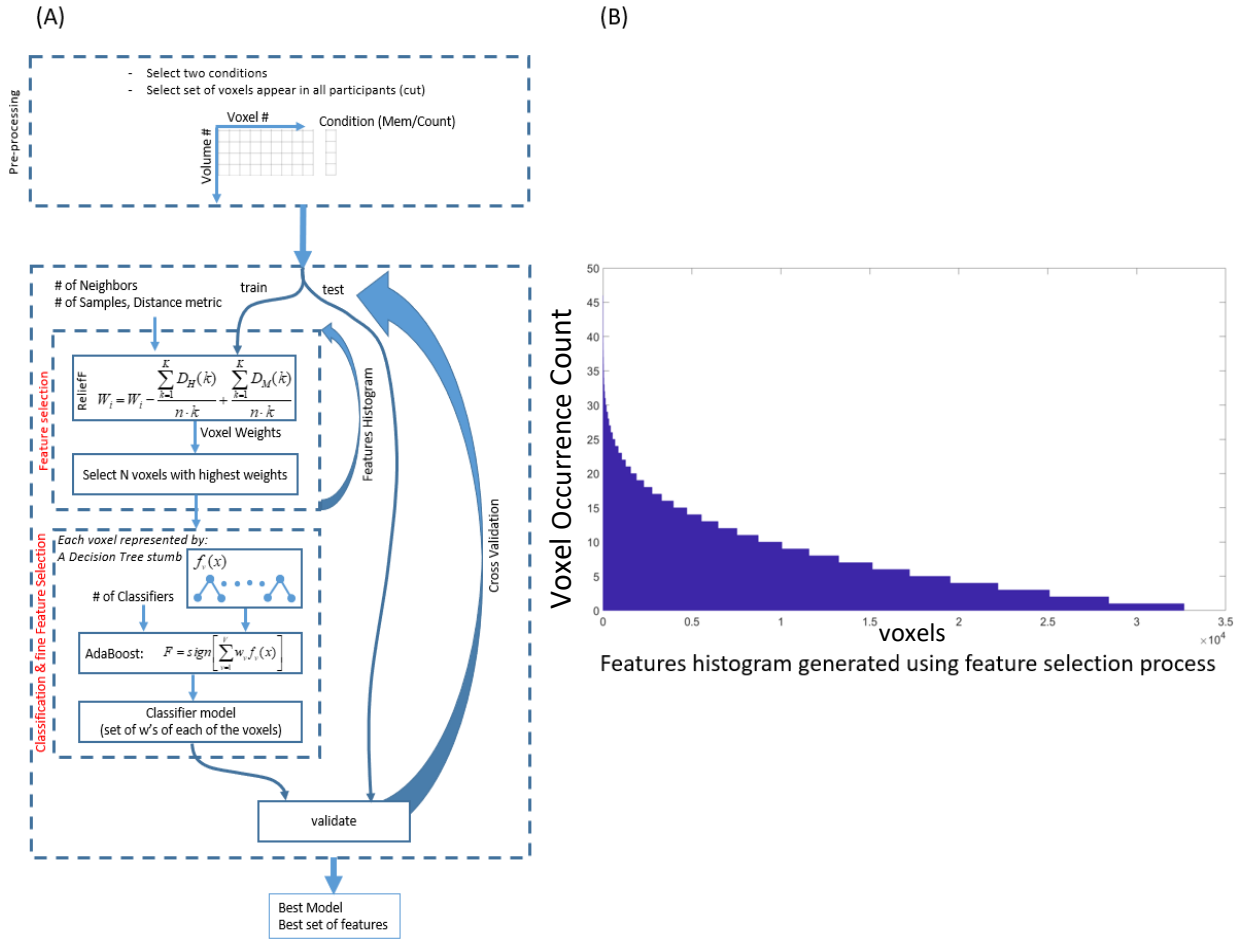

*Figure 2 – a) A sketch of the proposed method, that includes feature selection, classifier's training and testing phases, and b) an example of a histogram showing the frequency of specific voxels in the normalized brain coordinates being chosen, sorted by frequency during the feature selection process.*

### 2.2.4. Further generalization

We would like, if possible, to not only succeed in the classification but also try to identify the most important brain areas involved in this task. This is not direct from the classification for various reasons: (i) there is a variability in the selected features due to the cross-validation iterations that leaves out substantial amount of data (for validation), (ii) the methodology we use for the voxel feature selection process is actually non-deterministic, in the sense that there is randomness in the ReliefF algorithm action (the feature weighting algorithm) due to the sampling process (see section 2.2) and (iii) because the data may contain noise of non-physiological nature, for instance, task engagement may vary across scans.

To overcome some of these issues, we used a "histogram methodology". That is, we rerun the analysis (the left path of the diagram in Figure 2a) many times and give some indication of the frequency of a specific voxel being selected as feature on which classification algorithm will be trained. Initially, this was suggested in the work Boehm and Manevitz [27], but the computational resources were not available for that study at that time. An example of such a run can be seen in Figure 2b, where the x-axis indicate selected voxels (a.k.a. features) and the y-axis indicate frequency (i.e., the repeatability of a specific feature through different selections). From Figure 2b it can be seen that there is a relatively small set of voxels that is frequently repeated between the different folds. Using this histogram methodology (i) made the feature selection more stable and as a result substantially lowered the standard deviations as given in table 2 below, and (ii) supports our assumption that significant amount of voxels introduced noise into the "whole brain analysis" based classification scheme used in previous research.

Since the ReliefF is an iterative algorithm (i.e. for each feature (40,000 voxels) all data points (~600 scans per participant x 11 participants) are sampled over the amount of neighbors (4)) the algorithm is very time-consuming. Consequently sampling methodology was used; i.e., not all the data-points were used in order to evaluate a feature's influence at each fold). Besides, the decrease in running time, this method reduced the influence of noisy data-samples due to the sampling process. The sampling ratio of 10% was chosen as a reasonable compromise between the running time and representation of the dataset.

It is possible that further research on optimizing the parameters, such as sampling percentage, might improve these results presented in Section 3. Note that the main result is the *success* of separation by valence, which was not successfully generalized in Nawa [15].

## 3. Results

We report here three types of results; first, classification results for (i) each participant (i.e. within-participant) and (ii) cross-participants. Both analyses were made at the brain volume level. Second, we show that it is possible to correlate between the classification success and each user's reported degree of vividness experienced during the retrieval of memories. Lastly, a visual analysis of the voxels selected by the algorithm is presented. Note that due to the equal number of positive and negative memory tasks no balancing of the training dataset was required.

For both sets of classification results, in AdaBoost we stopped after choosing the best 260 voxels. The stoppage at this time was done after limited preliminary results (e.g., around 100 was substantially worse results, while even at the level of 600 there was no significant increase). Recall that the candidate 2500 voxels were chosen by their *univariate* quality.

***Within-participant results.*** For each participant, the recorded data was divided into train and test groups in the task block level, i.e. if a specific task is selected to the testing set, all of its 16 volume scans were excluded from the training set. Since there is a bold recovery period followed by execution of a neutral task, "block level" selection should be sufficient in order to avoid a data leakage between training and testing groups. The division is made randomly using 70%-30% to train and test group respectively (i.e., Monte Carlo cross validation) with 100 cross validation cycles. The ReliefF algorithm narrowed the data to 2500 voxels (approximately 6% of the voxels with highest weights were selected), and the final classification performed using a combination of 260 voxels selected from this subset using the AdaBoost algorithm as described in previous section.

The overall classification results for each of the 11 participants are presented in Table 1. A permutation test of independence was applied in order to validate the results. In order to calculate the p-value, 1000 permutations were performed at the block level per participant, resulting p-values of the classification result reported in the Table 1 below. (A p-value of 0.001 is the minimum in this case).

*Table 1 – Classification results within participant at task level. First column indicates participant number, second column is the average classification result and third is a result of a permutation test (1000 permutations used).*

| # | Classification Result | P-value derived from permutation tests |
|---|---|---|
| 1 | 0.852 | 0.032 |
| 2 | 0.684 | 0.045 |
| 3 | 0.750 | 0.030 |
| 4 | 0.869 | 0.028 |
| 5 | 0.617 | 0.041 |
| 6 | 0.782 | 0.021 |
| 7 | 0.649 | 0.039 |
| 8 | 0.924 | 0.008 |
| 9 | 0.897 | 0.015 |
| 10 | 0.951 | 0.003 |
| 11 | 0.971 | 0.003 |

*Cross-participant results.* Table 2 presents cross participants results. The cross-validation division into train and validation sets was 70% - 30% accordingly at the participant level (i.e., all the volumes of selected participants were included or excluded from the training set). The presented results were generated using 20 cross validation cycles, with random choice of the train/test groups. To validate the results, a permutation test was carried out, wherein we repeated the same classification process with permuted labels at block level. We repeated the test 100 times (a smaller number of permutations was used due to computational constraints) in order to create the probability function for randomized results and measure the significance level of the classification, which is p=0.039.

*Table 2 – Confusion matrix for cross-participants classification results. The number in the upper left corner is the average of the diagonal (i.e., the average classification result).*

| Cross Participants (0.624) | Mem Neg' | Mem Pos' |
|---|---|---|
| **Mem Neg'** | 0.633±0.018 | 0.367±0.018 |
| **Mem Pos'** | 0.384±0.024 | 0.616±0.024 |

*Correlating neuroimaging data with psychological data.* Figure 3 presents the correlation between the classification success rate (y-axis), within participant, and the reported degree of vividness (x axis) per each participant. Participants were asked to evaluate the vividness of the retrieved memories after scanning using a 11-point scale (0: low, 10: high) [15].

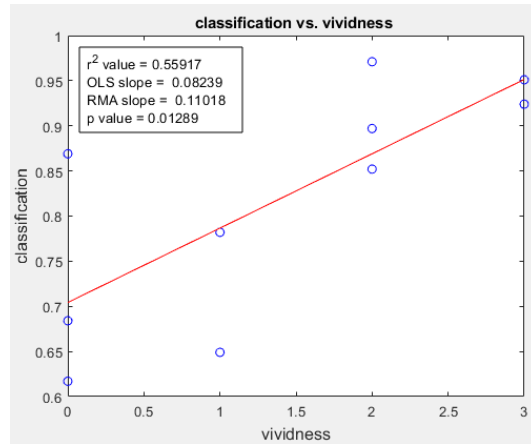

*Figure 3 – Correlation between the classification success rate and the reported degree of vividness. (Note, that this graph omits one outlier)*

### *Selected voxels visual analysis.*

For visual analysis of the brain areas which give the candidate voxels (i.e., the univariate choices) from which voxels were selected for the classification, the following procedure was performed: First, the feature selection step from the classification pipeline is executed using 100 folds for each participant, i.e., re-choosing the 70-30 training-validation split. During each fold, 2500 voxels with highest weights were selected and the occurrence of each voxel across the different folds was counted. Then, the resulting "voxel counts" from all the participants were superimposed (added) on the same MNI brain volume. Later, for visualization purposes, the resulting image was smoothed using a Gaussian kernel of 6-mm full width at half maximum and then manually thresholded to show the ~2500 voxels with the highest values (out of which clusters containing less than 5 voxels were removed). Figure 4 presents a view of the results of this procedure where warmer colors indicate higher values of the smoothed averaged histogram values. (This method eliminates isolated voxels even with relatively high histogram values; thereby allowing visualization of the most relevant brain areas based on our classification methods.)

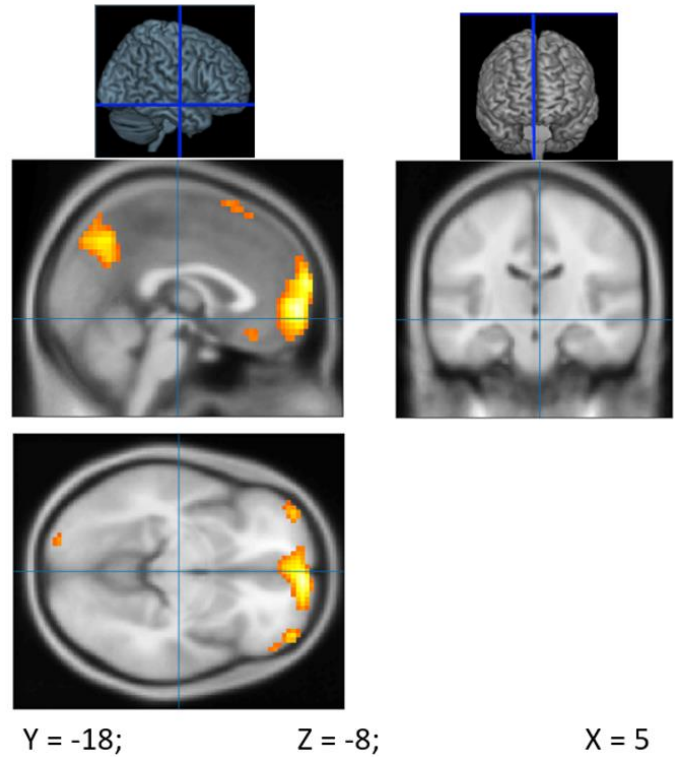

*Figure 4 – A representation (over a normalized MNI brain) of the average of the individual histograms (over the 100 feature selection folds) of best features for classification within individuals. Warmer colors indicate higher average histogram values. The two biggest clusters include the Precuneus (PCUN) and Superior Frontal Gyrus (SFG).*

It can be seen that there are two main (i.e., largest) clusters, located at the general area of Precuneus (PCUN) and Superior Frontal Gyrus (SFG), which are known to be related to recollection of emotional encodings (SFG) and with episodic memories (Precuneus).

## 4. Summary and Discussion

In Nawa and Ando [15] the researchers felt that it was preferable to do a whole-brain analysis; and they suggested some potential machine learning advantages. Most important of these was the suggestion that since the autobiography memory task is a complex one; much more of the brain (and hence its activation pattern) would be involved in the task; and therefore, using all of the voxels would boost the percentage of the appropriate signal in the data being used for the classification. We do not feel the work in this paper contradicts this and, in fact, it is reasonably understood that memory tasks are quite sophisticated cognitively and apparently utilize many brain regions. (For example, frontal brain structures involved in working memory also underlie declarative memory in both encoding and recall [28].)

Nonetheless, in this paper we clearly saw that the use of extensive feature selection (in this case using ReliefF and AdaBoost) opens the door to other techniques of machine learning that allows us to successfully classify the cross-participant autobiography positive versus negative valence, something at which the techniques of [15] did not succeed. Furthermore, in the tasks at which they did succeed, the feature selection methods in this paper gave much more accurate results as can be seen by comparing our results with those appearing in Table 1 and Figure 2 in [15]. Moreover, while the analysis in [15] gave significant results both cross-participant and within participant for Countdown versus Autobiography; the feature selection methods used here gave much more accurate results (not reported here). (Note that applying a feature selection step before the machine learning training is not the same as an a priori selection of anatomically or functionally defined regions of interest (ROIs). Instead, we used the strength of the machine learning classification tools themselves to properly select individual voxels directly.)

We believe it is clear that instead of using "full brains", feature selection allows us to do more accurate and deal with more delicate separation tasks.

In our opinion, the reason for this is not the focusing on areas responsible for the tasks; but rather on areas having a higher signal to noise ratio; where the "signal" is information related to the task. In principle, such voxels do not necessarily indicate that they are actually *causally involved* in the task; just that from the information perspective, there is a correlation which can be levered for classification.

Accordingly, one has to be quite careful in interpreting diagrams (like Figure 4). We are only discovering *correlates.* Looked at in this way; we might find that, e.g., if a subject was going to blink an eye (right or left) based on positive or negative valence; a simple voxel in the motor cortex would probably be sufficient to clearly distinguish between the cases; even though the voxel had nothing to do with the memory recall. Since our tools for classifying are mostly correlation based, it may be that the more focused the features are, the better results we can expect.

There are several directions that might be followed up from this work: (i) it seems to us that "significance vs non-significance" as a measure can be substantially refined by looking at the degree of classification. For example, in the cases where Nawa and Ando ([15]) succeeded; we did as well, but to a much higher (around 15% difference) percentage of classification accuracy. (ii) We now have a clear view as to why more advanced feature selections are advantageous (e.g., compared to ANOVA and certainly to no feature selection).

We point out that the use of multivariate methods can in principle find non-local inter-relationships between features. In this work, especially that described in section 3, tends in fact to eliminate this potential. AdaBoost in principle searches at each stage for a voxel that would add the most information, which gives a bias against nearby voxel who often carry similar information. (Philosophically, it has a similar intuition to what is called *"active learning"* in AI [29]; or *"optimal experimental design"* [30], [31] in the statistics literature.) On the other hand, the possible candidates for selection in AdaBoost are chosen univariately, as the voxels, which *by themselves* cause the best separation of the classes. This multi-variate choice of voxels is what allows good classification from only 260 voxels. Potentially the inter-action between such chosen voxels might offer further biological insight. We hope that future work will more deeply consider this aspect of the multi-variate methodology.

In section 3, on the other hand, we used a histogram approach to locate the main areas in the brain, leveraging the biological variance in each scan; and then following with blurring and thresholding for visualization purposes.

In summary, the take-home message is that subtle cognitive tasks can be classified using a combination of machine learning techniques; more specifically, the current results suggest that the differences in terms of neurophysiological mechanisms characterizing these processes are sufficiently coherent across participants, thus enabling above-chance cross-participant classification.

# Acknowledgements

The CiNet Research Institute hosted LM (twice) and AF (once) to work with NEN on this work. We thank its staff and especially Director Takahisa Taguchi for their support and encouragement of this work. This work was performed under an MOU between CiNet and the University of Haifa. Some of the computations were performed under an equipment grant of NVIDIA corporation to the Neurocomputation Laboratory at the Cesarea Research Institute, University of Haifa. NEN was partially supported by the Japan Society for the Promotion of Science under a Grant-in-Aid for Scientific Research (JPS KAKENHI grant number JP17K00220).